\documentclass[sn-mathphys,Numbered]{sn-jnl}

\usepackage{graphicx}%
\usepackage{multirow}%
\usepackage{amsmath,amssymb,amsfonts}%
\usepackage{amssymb, amsmath}
\usepackage{amsthm}%
\usepackage{mathrsfs}%
\usepackage[title]{appendix}%
\usepackage{xcolor}%
\usepackage{textcomp}%
\usepackage{manyfoot}%
\usepackage{booktabs}%
\usepackage{algorithm}%
\usepackage{algorithmicx}%
\usepackage{algpseudocode}%
\usepackage{listings}%
\usepackage{bm}
\usepackage{times}

\theoremstyle{thmstyleone}%
\theoremstyle{thmstyletwo}%

\theoremstyle{thmstylethree}%

\begin{document}

\title[Modelling wildland fire burn severity in California using a spatial super learner approach]{Modelling wildland fire burn severity in California using a spatial Super Learner approach}


\author[1]{\fnm{Nicholas} \sur{ Simafranca}}\email{simaf001@umn.edu}
\equalcont{These authors contributed equally to this work.}

\author[2]{\fnm{Bryant } \sur{Willoughby}}\email{btwillou@ncsu.edu}
\equalcont{These authors contributed equally to this work.}

\author[3]{\fnm{Erin} \sur{O'Neil}}\email{erinponeil@g.ucla.edu}
\equalcont{These authors contributed equally to this work.}

\author[4]{\fnm{Sophie} \sur{Farr}}\email{sfarr@vassar.edu}
\equalcont{These authors contributed equally to this work.}

\author[2]{\fnm{Brian J} \sur{Reich}}\email{bjreich@ncsu.edu}

\author[2]{\fnm{Naomi } \sur{Giertych}}\email{ngierty@ncsu.edu}

\author[5]{\fnm{Margaret} \sur{ Johnson}}\email{maggie.johnson@jpl.nasa.gov}

\author[5]{\fnm{Madeleine} \sur{Pascolini-Campbell}}\email{madeleine.a.pascolini-campbell@jpl.nasa.gov}

\affil[1]{
\orgname{University of Minnesota - Twin Cities}}

\affil[2]{
\orgname{North Carolina State University}}

\affil[3]{
\orgname{University of California - Los Angeles}}

\affil[4]{
\orgname{Vassar College}} 

\affil[5]{
\orgname{Jet Propulsion Laboratory, California Institute of Technology, Pasadena, CA}}

\abstract{Given the increasing prevalence of wildland fires in the Western US, there is a critical need to develop tools to understand and accurately predict burn severity. We develop a  machine learning model to predict post-fire burn severity using pre-fire remotely sensed data. Hydrological, ecological, and topographical variables collected from four regions of California — the sites of the Kincade fire (2019), the CZU Lightning Complex fire (2020), the Windy fire (2021), and the KNP Fire (2021) — are used as predictors of the differenced normalized burn ratio. We hypothesize that a Super Learner (SL) algorithm that accounts for spatial autocorrelation using Vecchia’s Gaussian approximation will accurately model burn severity. In all combinations of test and training sets explored, the results of our model showed the SL algorithm outperformed standard Linear Regression methods. After fitting and verifying the performance of the SL model, we use interpretable machine learning tools to determine the main drivers of severe burn damage, including greenness, elevation and fire weather variables. These findings provide actionable insights that enable communities to strategize interventions, such as early fire detection systems, pre-fire season vegetation clearing activities, and resource allocation during emergency responses. When implemented, this model has the potential to minimize the loss of human life, property, resources, and ecosystems in California. }




\maketitle

\section{Introduction}\label{s:intro}

Current trends reveal an increase in the magnitude and frequency of wildland fires in the Western US \cite[]{Dennison}. The rise of ``megafires'' in recent years pose a threat to the environment, human life, property, and resources \cite[]{Coen}. Wildland fires disturb microclimates by causing ``type conversions'' of landscapes \cite[]{Coop} where once dominate vegetation (such as forestry) is replaced with new vegetation (such as grasses). ``Type conversions'' intensify the conditions conducive to wildland fires by creating a hotter landscape with less moisture \cite[]{Coop}. Not only does this pose an ecological threat, but it also has dangerous social implications. For one, wildland fire smoke has proven to be detrimental to respiratory health \cite[]{Heaney}. Furthermore, communities bear the economic burden of recovering from devastating structural damages. Therefore, studying the factors that influence the severity and spread of burn incidents has reached a critical point. By understanding the drivers of wildland fires, communities can effectively allocate their time and resources to minimize losses. 

Burn severity is a metric used to evaluate the post-fire damage to soil and vegetation \cite[]{Keeley}. This metric is influenced by the availability and flammability of fuels, environmental stressors, and topography \cite[]{Coen}. Previous studies have shown that plant water stress variables are important for predicting burn severity \cite[]{Pascolini-Cambell2021ECOSTRESS}. There is also evidence that pre-season soil moisture is a strong predictor \cite[]{Jensen}. Data from the ECOsystem and Spaceborne Thermal Radiometer Experiment on Space Station (ECOSTRESS) satellite \citep{fisher2020ecostress} provides high spatial and temporal resolution of hydrological information such as evapotranspiration (ET), evaporative stress index (ESI), and water use efficiency (WUE). We incorporate these variables into our study due to their relevance in characterizing fuel amount and flammability. We hypothesize that these variables will increase the predictability of burn severity. Regions with greater plant productivity, and more fire fuel, tend to have higher measures of ET \cite[]{Fisher}. Moreover, ESI, which evaluates the moisture available to vegetation, serves as an indicator of the flammability of those fuels \cite[]{Huang}. Finally, WUE measurements provide insight to the vulnerability of plants to climatic stressors \cite[]{Pascolini-Cambell2021ECOSTRESS}. The complex relationship between these factors influence the spatial patterns of burn severity \citep{Kane_2015}. Due to this complexity, the prediction of the spatial patterns of burn severity has garnered the attention of researchers in recent decades.

Traditional physics-based simulations that capture physical processes \citep{Hoffman} pose challenges. These simulators are time dependent and rely on numerical solutions to computational fluid dynamics. Jain \cite[]{Jain} reports that it is unfeasible to apply such models on large scales and often produce low accuracy. To overcome these issues, many wildland fire researchers turn to empirical and statistical models. However, this requires modelling nonlinear relationships, introducing complexity. Machine learning (ML) has emerged as a contemporary approach in wildland fire research. This approach is independent of the implementation of physical processes. Instead, it learns directly from data \cite[]{Jain}. Many ML methods, henceforth referred to as ``base learners," have been applied to wildland fire prediction including Random Forests (RF), Maximum Entropy, Artificial Neural Networks, Decision Trees, Support Vector Machines (SVM), and Genetic Algorithms \cite[]{Jain}. 

A study of the Basic Complex fire (2008) in Big-Sur, California compared the performance of RF, Gaussian Process Regression (GPR), and SVM to multiple regression in assessing burn severity. RF performed the best and reduced model error by 48\% compared to Linear Regression. This reduction in model error is likely due to its ensemble learning approach \cite[]{Hultquist}. Ensemble learning merges multiple base learners to solve a single learning problem \cite[]{Zhou_2021}. To date, the use of ensemble ML models to predict burn severity remains relatively unexplored. Current literature indicates that ensemble learning outperforms individual learners \citep[e.g.,][]{Van}. When base algorithms are accurate and diverse, performance is further enhanced \cite[]{Zhou_2021}. Our model will incorporate a Super Learner (SL) ensemble algorithm \citep{vanderLaan2007SuperLearner}, which aggregates diverse base learners, to forecast wildland burn severity. In addition, our model will account for spatial autocorrelation. To the best of our knowledge, this has yet to be done for predicting burn severity. 

Our motivation for combining ensemble prediction and spatial modeling stems from current literature which reveals the effectiveness of combining ML algorithms with traditional spatial statistic methods. When handling spatially-dependent data, ML algorithms alone are limited in their ability to account for common errors that arise in geostatistics such as data gaps \citep{Wikle}. Furthermore, unlike geostatistical methods, many common machine learning algorithms are not able to provide estimates of prediction or classification error \citep{Wikle}. 

Applying a ML algorithm in conjunction with a statistical method is likely to be more applicable to spatially-correlated data and yield stronger predictions. Current literature reveal spatial prediction is used in tandem with diverse ML algorithms. There are multiple ways to implement a spatial ML model; one way is to account for spatial correlation after implementing the ML algorithm \cite[]{enviromentspatialRF}. Another is attempting to account for spatial correlation within the ML algorithm, such as Random Forest \cite[]{RFspatial}. Li \cite{enviromentspatialML} found that a combination of Random Forest and Ordinary Kriging (a form of spatial prediction) provided the best results out of 23 methods.  \cite{NeuralKriging} got the most accurate predictions at new locations after training with a Multi-layer Neural Network that includes spatial relationships compared to just ordinal Kriging and normal Multi-layer Neural Network. \cite[]{NeuralKrigingIRAN} also found benefits from combing Neural Networks and Kriging for spatial prediction. 

A mixed spatial ML model was proposed by \cite{RFGLS} that incorporates a RF machine learning algorithm and accounts for spatial autocorrelation using a Gaussian processes. The mixed-model approach proved superior to RF alone when evaluating spatially correlated data. Davies \cite{EnsemblespatialStats} examines the use of an ensemble ML algorithm that includes Kriging in its base learner library. The favorabilty of a mixed-method approach was further confirmed by their results which found that SL performs as good or better than the best base learner in spatial prediction. Another study found that optimizing the predictions from both Kriging and SL separately provided accurate estimation of geological attributes \cite[]{SLKriging}. Because there is strong geospatial correlation in our wildland fire datasets, we expect to find that accounting for spatial correlation on the ensemble predictions will enhance the predictability of our SL model. 

In this study, we investigate the spatial pattern of burn severity of four wildfires between 2019-2021. Two fires, the Kincade and San Mateo–Santa Cruz Unit (CZU) Lightning Complex fire, represent fires occurring along the Northern Pacific Coast in California \citep{Calfire}. This territory is characterized by a warm/hot summer Mediterranean climate. In contrast, the Windy  and Sequoia and Kings Canyon National Park (KNP) Complex represent fires that occur in the Sierra Nevada region which spans the majority of the inland mid-latitudes of California 
\citep{InciWeb_2022}. A dry summer subarctic climate can be found in this region. This paper is motivated by the questions:
\begin{itemize}
    \item How can we relate areas of intense burn to vegetation, weather, and topography?
    \item Can a Super Learner algorithm accurately predict post-wildfire burn severity throughout California? 
    \item Which variables are most important to burn severity? 
\end{itemize}

The remainder of the paper proceeds as follows.  The wildland fire data are described in Section \ref{s:data}.  Section \ref{s:method} introduces the methods and computational algorithms.  The results are summarized in Section \ref{s:results}.  Section \ref{s:discussion} concludes. 

\section{Data description}\label{s:data}

\begin{figure}
\centering 
\includegraphics[width=0.75\textwidth]{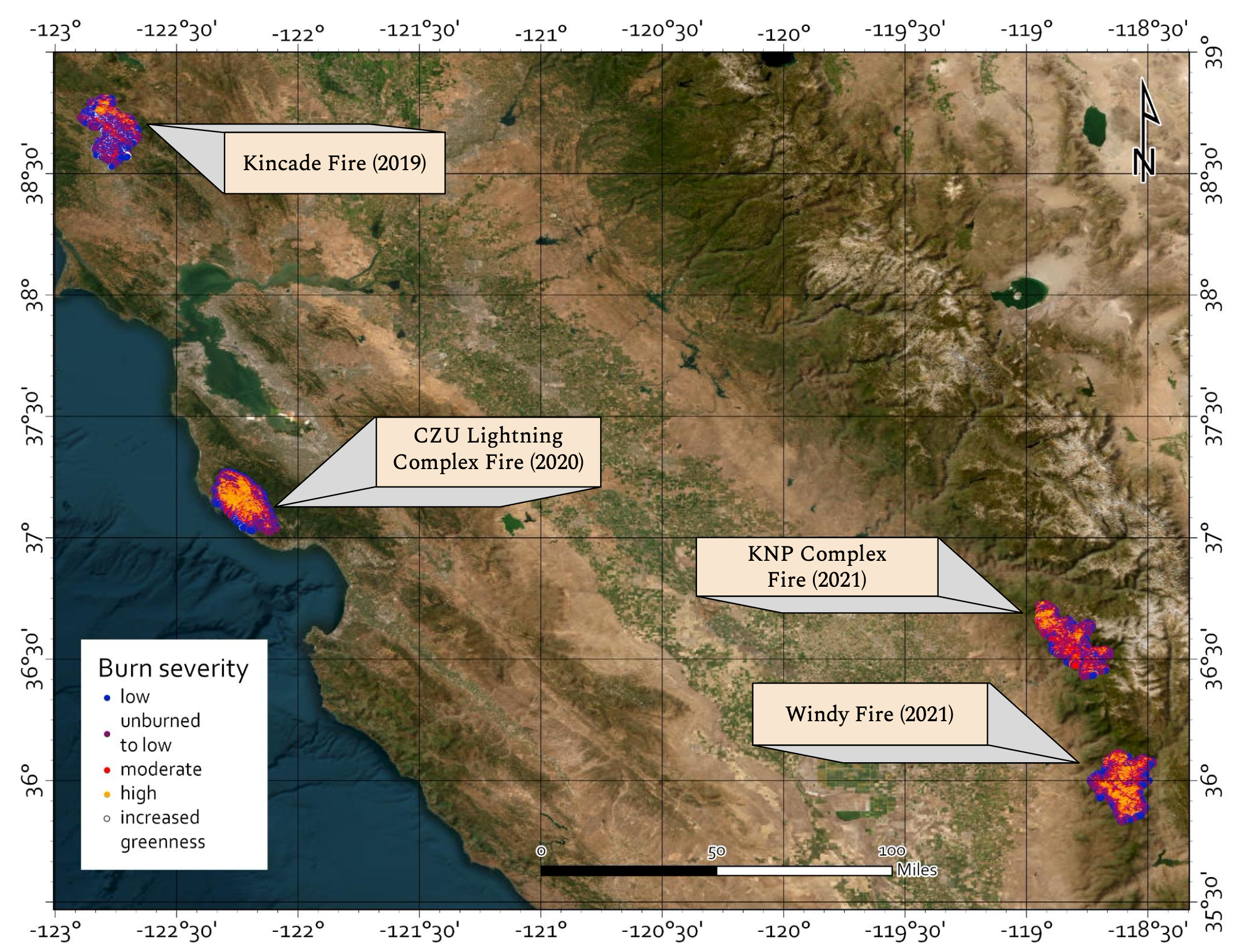}
\caption{Maps of the burn severity measure (dNBR) for the Californian wildland fires analyzed in the study.}\label{f:fires}
\end{figure}

We analyze data from four fires in California: the Kincade fire, the CZU Lightning Complex fire, the Windy fire, and the KNP Fire. The fires are described in Table \ref{tab:fires} and plotted in Figure \ref{f:fires}. The data sources described in this section include the ECOSTRESS satellite, Harmonized Landsat Sentinel, Digital Elevation Model, Moderate Resolution Imaging Spectroradiometer (MODIS), Panchromatic Remote-sensing Instrument for Stereo Mapping (PRISM), and Soil Moisture Active Passive (SMAP) (Table \ref{tab:covariates}). All products were regridded to 70m resolution. The burn severity response variable (Section \ref{ss:burn_severity}) is obtained after the fire, and thus is observed only once per pixel per fire. The predictor variables (Sections \ref{ss:others}) are all measured before the onset of the fire and some are measured multiple times in the two weeks prior to the onset of the fire. For instance, for the KNP fire, the Water Use Efficiency (WUE) variable was measured four times before the onset of ignition (Aug. 26, Aug. 30, Sep. 03, and Sep. 07). In Section \ref{ss:manipulation} we describe how we handle missing data and how we resolve the repeated measures. 

\begin{table}
\small
\label{tab:fires}
\begin{tabular}{ll|ccccc}
 Fire & Location & Ignited & Fully Contained & Acreage & \# of Pixels \\
 \hline
Kincade & Sonoma County, CA & Oct. 23, 2019 & Nov. 6, 2019 & 77,758 & 82,125   \\
CZU & Santa Cruz \& San Mateo, CA & Aug. 16, 2020 & Sept. 22, 2020 & 86,509 & 88,581   \\
Windy & Sierra Nevada, CA & Sept. 9, 2021 & Nov. 15, 2021 & 97,528 & 99,458  \\
KNP & Sierra Nevada, CA & Sept. 9, 2021 & Dec. 16, 2021 & 88,307 & 92,171   
\end{tabular}
\caption{Summary of wildland fires used in the study.}
\end{table}

\subsection{Burn severity}\label{ss:burn_severity}
Our outcome variable of interest is burn severity, which is a function of the normalized burn ratio (NBR).  NBR compares measurements in the near infrared (NIR) and the short-wave infrared (SWIR) sections of the electromagnetic spectrum as
\begin{equation}\label{e:NBR}
    NBR = \frac{NIR - SWIR}{NIR + SWIR}.
\end{equation}
Healthy vegetation reflects in the NIR section whereas burned vegetation reflects stronger in the SWIR section
\citep{Pascolini-Cambell2021ECOSTRESS}.
The response variable is the differenced normalized burn ratio, or dNBR, which is the difference between pre-fire NBR and post-fire NBR so that a large dNBR value implies extreme burn severity and vice versa. Landsat 30m images were used to calculate dNBR values. The dNBR data was retrieved from the interagency program ``Monitoring Trends in Burn Severity (MTBS)" and scaled by a factor of 1,000. This continuous measure of burn severity can be reduced to a categorical measure using the following categories: high enhanced growth (dNBR from -500 -- -251), low enhanced regrowth (-250 -- -101), unburned (-100 -- 99), low (100 -- 269), moderate-low (270 -- 439), moderate-high (440 -- 659), and high (660 -- 1300). We build our model for the continuous severity response, but evaluate performance for correctly predicting the categorical measure.  

\subsection{Covariates}\label{ss:others}
 Data for each covariate was collected within a two-week time period preceding fire onset. We acquired variables at a 70m resolution describing plant water stress from the ECOSTRESS \citep{fisher2020ecostress}, which gives information on fuel availability, drought, and other potentially important predictors of wildland fire burn severity \cite[]{Pascolini-Cambell2021ECOSTRESS}. Evapotranspiration (ET) measures the amount of water being lost in the soil from both evaporation from soil surface and transpiration from the plant leaves. The Evaportaive Stress Index (ESI) is the ratio of actual ET to potential ET, and can be an indicator of drought. Water Use Efficiency (WUE) indicates how a plant responds to stress, such as short-term drought. Land Surface Temperature (LST) is defined as the temperature that the land would feel to the touch. 
 
 In addition to ECOSTRESS measurements, we include variables describing vegetation. The Normalized Differenced Vegetation Index (NDVI) and the Leaf Area Index (LAI) quantifies the plant canopies. The morning soil moisture (AM SM) and afternoon soil moisture (PM SM) were also variables of interest.  Moreover, we have variables describing fire weather two weeks before the fire. These include both morning and evening soil moisture, daily average dew point temperature (TDMEAN), daily average air temperature (TMEAN), vapor pressure maximum (VPDMAX), and vapor pressure minimum (VPDMIN). Lastly, the variables elevation, slope, and aspect describe the topography of the region. A summary of the covariates with their corresponding satellite origin and original resolutions are provided in Table \ref{tab:covariates}.

\begin{table}
\centering
\begin{tabular}{l|cc}
 Variable  & Source & Resolution\\
 \hline
 Evapotranspiration (ET)            & ECOSTRESS    & 70m  \\
 Evaportaive Stress Index (ESI)        & ECOSTRESS    & 70m  \\
 Water Use Efficiency (WUE)   & ECOSTRESS    & 70m  \\
 Land Surface Temperature (LST)    & ECOSTRESS    & 70m  \\
 Normalized Differened Vegetation Index (NDVI)   & Harmonized Landsat Sentinel    &30m  \\
 Leaf Area Index (LAI) & MODIS  & 500m   \\
 AM Soil Moisture (AM SM) & SMAP & 9km \\
 PM Soil Moisture (PM SM) & SMAP & 9km \\
 Daily Average Dew Point Temperature (TDMEAN) & PRISM  & 4km  \\
 Daily Average Air Temperature (TMEAN) & PRISM  & 4km  \\
 Vapor Pressure Deficit Maximum (VPDMAX) & PRISM  & 4km\\
 Vapor Pressure Deficit Minimum (VPDMIN) & PRISM  & 4km\\
 Elevation & Digital Elevation Model & 30m\\
 Aspect    &Digital Elevation Model &30m\\
 Slope&   Digital Elevation Model   & 30m
\end{tabular}
\caption{Variables used as covariates in the Super Learner model.  With the exception of the final three variables, all variable were included as both their current state and temporal trend.}\label{tab:covariates}
\end{table}

\subsection{Data Manipulation}\label{ss:manipulation}

The variables used in the study are derived from different sources and thus have different spatial resolutions (Table \ref{tab:covariates}). This change of support is resolved by converting all variables to the ECOSTRESS 70m grid using area-weighted averaging.  We used bilinear interpolation for continuous variables and nearest neighbor for categorical variables using a raster package in R \citep{Hijmans_2023b}.  Missing data is present due to factors such as cloud coverage.  We removed variables with more than 28\% missing pixels across each geographic region. For the remaining variables, missing observations were imputed using K Nearest Neighbor imputation, with $K=10$ and distance defined using latitude and longitude.

Many of the variables are collected at different times within the two-week period leading up to the fire ignition.  To resolve collinearly between subsequent measurements and harmonize variables across fires with a different number of replications of the variables, we convert the sequence of observations to estimates of the current value and the trend at the time of ignition for each pixel. Variables that are treated in this way will be labeled with the term ``current" or ``trend." For instance, ``TDMEAN (trend)'' refers to the change in mean dew point temperature over the time interval of our data and``TDMEAN (current)" refers to the mean dew point temperature at the onset of the fire. Estimates are made using Linear Regression separately by variable and pixel. For example, let $ET_{it}$   be the observed evapotranspiration (ET) at spatial location $\textbf{s}_i$ $t$  days before ignition.  We then fit the linear model $\mbox{E}(ET_{it}) = \beta_{0i} + \beta_{1i}t$ and use the least-squares estimates of $\beta_{0i}$ and $\beta_{1i}$ as covariates to summarize the current value and trend in ET, respectively, for pixel $i$ at the time of ignition.  

\section{Statistical methods}\label{s:method}
\begin{figure}
\centering 
\includegraphics[width=1\textwidth]{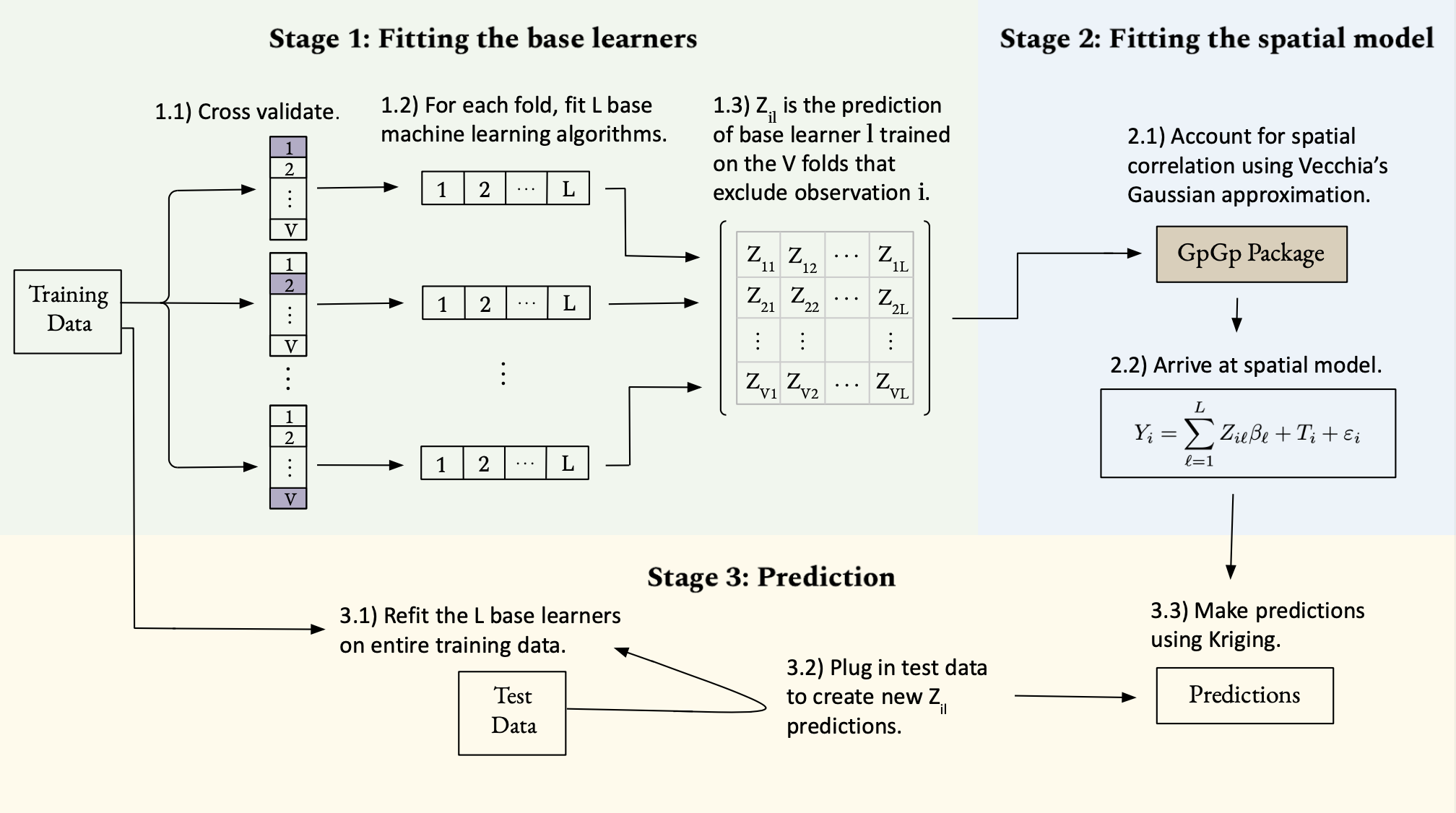}
\caption{Flowchart depicting Super Learner algorithm (based on diagram from \cite{vanderLaan2007SuperLearner})} \label{f:diagram}
\end{figure}

In this section, we discuss the methods used for predicting the continuous burn severity response variable; the method is summarized in Figure \ref{f:diagram}. We implement a spatial extension of a SL  algorithm, an ensemble learning method that combines base learners to achieve superior predictive accuracy. 
Section \ref{ss:spatialmodel} gives the overall model framework.  The model is fit in two stages: in the first stage (Section \ref{ss:superlearner1}) we use non-spatial regression to train the base learners and in the second stage (Section \ref{ss:superlearner2}) we use the first-stage estimates as covariates and fit a spatial process model used for prediction (Section \ref{ss:prediction}).  

\subsection{Spatial Super Learner model}\label{ss:spatialmodel}
Let $Y_i$ be the continuous measure of burn severity (dNBR) at spatial location $\textbf{s}_i$, and $\textbf{X}_i = (X_{i1}, ..., X_{ip})$ be the associated covariates (the $p=27$ variables in Table~\ref{tab:covariates}). Rather than using the covariates directly in the linear model, we pass them through $L$ machine learning algorithms (e.g., RF) to allow for interactions and non-linearity. Denote $Z_{il} = f_l(\textbf{X}_i)$ as the non-linear function of the covariates determined by learner $l\in\{1,...,L\}$. The spatial model is 
\begin{equation}\label{e:Ymodel}
   Y_i = \beta_0 + \sum_{l=1}^LZ_{il}\beta_l + T_i + \varepsilon_i,
\end{equation}
where $\beta_0$ is the intercept, $\beta_l$ is the weight given to learner $l$, $T_i$ is spatially-correlated error and $\varepsilon_i\overset{\text{idd}}{\sim}{Normal}(0,\tau^2)$ is independent error that accounts for small-scale, unexplained variance. The spatial error term $T_i$ captures correlation not explained by the covariates and is taken to be a stationary Gaussian process with E$(T_i)=0$, Var$(T_i)=\sigma^2$ and isotropic exponential correlation function Cor$(T_i,T_j) = \exp(-d_{ij}/\phi)$, with $d_{ij}$ denoting the distance between $\textbf{s}_i$ and $\textbf{s}_j$ and $\phi$ is the spatial range parameter.

\subsection{Stage 1: Fitting the base learners}\label{ss:superlearner1}

The Super Learner is a stacking ensemble method, introduced by \cite{vanderLaan2007SuperLearner}, that seeks the optimal combination of base learners ensuring predictive performance at least as well or better than the best performing base learner. The effectiveness of SL is attributed to the diversity of its base learners. Different learners, each offering a unique approach to problem solving, contribute to a robust and adaptive model. The choice of base learners can be customized based on the requirements or constraints of a given problem. We incorporate $L = 11$ diverse base learners: Elastic Net \citep[]{zou2005regularization}, Decision Tree Regression \citep[]{morgan1963problems}, Ridge Regression \citep[]{hoerl1970ridge1, hoerl1970ridge2}, Lasso Regression \citep[]{tibshirani1996regression}, K Nearest Neighbors Regression \citep[]{cover1967nearest}, Gradient Boosting Regression \citep[]{friedman2001greedy}, XGBoost Regression \citep[]{chen2016xgboost}, Bagging Regression \citep[]{breiman1996bagging}, Random Forest Regression \citep[]{breiman2001random}, Extra Trees Regression \citep[]{geurts2006extremely}, and Multilayered Perceptron Regression \citep[]{rosenblatt1958perceptron}. 

The base learners are fit individually and without regard to spatial correlation. For model-fitting (we take a slightly different approach for prediction in Section \ref{ss:prediction}), the training data are split into 10 folds using completely random sampling over $i$.  The covariate $Z_{il}$ in (\ref{e:Ymodel}) is the prediction of base learner $l$ trained on the 9 folds that exclude fold $i$. Cross validating tests our models ability to perform well on unseen data and prevents possible overfitting. This method of cross validation was implemented for both the within-fire (those predictions from which the training and test data are from the same fire) and combined fire datasets (those predictions from which the training data is derived of multiple fires and tested on a subset of that data). More details of base learner training are given in Appendix A.1. 

\subsection{Stage 2: Fitting the spatial model}\label{ss:superlearner2}

For the purpose of estimating the parameters $\{\beta_0,...,\beta_p,\sigma^2,\tau^2,\phi\}$ in (\ref{e:Ymodel}), we treat the base learner predictions $Z_{il}$ as fixed and known covariates and fit a standard spatial linear regression model.  Unlike some SL estimators \citep{vanderLaan2007SuperLearner}, we do not restrict the weights $\beta_1,...,\beta_L$ to be positive or sum to zero. Instead, we estimate the weights using uncontrained spatial regression to compensate for fitting the base learners using non-spatial regression.     

Due to the large number pixels, the exact maximum likelihood estimator cannot be computed and so we use the Vecchia approximation \citep{Vecchia_1988} implemented in the {\tt GpGp} package \citep{GpGp}.  The Vecchia approximation orders the observations and approximates the joint likelihood as a product of the univariate conditional distributions of each site given a local subset of sites that appear before the site in the ordering.  In our analysis, we use the default number of nearby sites in the subset ($k=15$) and ordering scheme.

\subsection{Stage 3: Prediction}\label{ss:prediction}

For prediction on the training data, we refit the base learners using the entire training dataset and compute fitted values at the training locations and predicted values at the testing locations as the covariates, $Z_{il}$ (Figure \ref{f:diagram}).  Given these covariates, standard Kriging predictions can be applied.  However, due to the size of the training data, standard methods are too computationally expensive and so we use the local Kriging prediction option in the {\tt GpGp} package that makes predictions at a test location based on the nearest $k$ training set observations.  Given the predicted values for the continuous measure of burn severity, we predict the categorical burn severity measure by discretizing the Kriging prediction into the corresponding category using the thresholds given in Section \ref{ss:burn_severity}. Details of fitting the spatial model and subsequent prediction are given in Appendix A.2. 

\section{Results}\label{s:results}
\subsection{Within-fire predictions}\label{s:app:compare1}

\begin{table}
\centering
\begin{tabular}{ll|l}
Mean & Covariance & Model\\\hline
LR  & Independent & $Y_i = \beta_0 + \sum_{p=1}^PX_{ip}\beta_p  + \varepsilon_i$ \\
LR & Spatial & $Y_i = \beta_0 + \sum_{p=1}^PX_{ip}\beta_p  + T_i + \varepsilon_i$  \\
SL & Independent & $Y_i = \beta_0 + \sum_{l=1}^LZ_{il}\beta_l + \varepsilon_i$ \\
SL & Spatial & $Y_i = \beta_0 + \sum_{l=1}^LZ_{il}\beta_l + T_i + \varepsilon_i$\\
&\vspace{-6pt}\\
\end{tabular}
\caption{Linear Regression (LR) versus Super Learner (SL) for the mean with spatial versus independent errors for $p\in\{1,...,P\}$ covariates $l\in\{1,...,L\}$ base learners. The models differ by whether the original covariates ($X_{ip}$) or base learners ($Z_{il}$) are included in the mean and whether the spatial random effects ($T_i$) are included in the covariance. }
\label{tab:models}
\end{table}

We first analyze the four fires separately. We fit four models (Table \ref{tab:models}): Linear Regression versus SL for the mean with spatial versus independent errors. The first step in the spatial SL method is to compute the base learner predictions, $Z_{il}$. Table \ref{tab:IX:base_correlation} gives the sample correlations between the base learners for the KNP fire.  While there are some highly correlated base learners, there are many with moderate correlation and so as desired the set of base learners is diverse and conducive to ensemble prediction.   Table \ref{tab:Fit:withinfire} gives estimates of the weight parameters, $\beta_l$, for the spatial SL fit for each fire.    For all fires, Extra Trees Regression has the most weight followed by Multilayer Perceptron Regression and XGBoost Regression.  The consistency of the results across fires suggests the predictive model may be generalizable across similar fires.

\begin{table}
\begin{tabular}{l| rrrrrrrrrr}
\rotatebox{90}{} & \rotatebox{90}{Decision Tree} & \rotatebox{90}{Ridge Reg} & \rotatebox{90}{Lasso Reg}    & \rotatebox{90}{KNeighbors Reg} & \rotatebox{90}{Gradient Boosting Reg} & \rotatebox{90}{XGBoost Reg } & \rotatebox{90}{Bagging Reg}& \rotatebox{90}{Random Forest Reg } & \rotatebox{90}{Extra Trees Reg} & \rotatebox{90}{Multilayered Perceptron Reg } \\\hline
            Elastic Net          & 0.39     & 0.86     & 0.94     & 0.48     & 0.50     & 0.46     & 0.49 & 0.49 & 0.48 & 0.47 \\
            Decision Tree     &          & 0.44     & 0.43     & 0.74     & 0.79     & 0.82     & 0.82 & 0.82 & 0.81 & 0.74 \\
            Ridge Reg     &      &         & 0.96     & 0.55     & 0.58     & 0.54     & 0.56 & 0.56 & 0.55 &0.55\\
            Lasso Reg     &      &      &          & 0.53     & 0.55     & 0.51     & 0.54 & 0.54 & 0.53 & 0.52\\
            KNeighbors Reg     &       &      &      &         & 0.90     & 0.89     & 0.91 & 0.91 & 0.92 & 0.86 \\
            Gradient Boosting Reg     &      &      &      &      &        &  0.95     &  0.97 & 0.97 & 0.86 &0.90 \\
            XGBoost Reg     &      &      &      &      &      &        & 0.97 & 0.97 & 0.97 & 0.89 \\
            Bagging Reg     &      &      &      &      &      &      &      & 0.99 & 0.98 & 0.91\\
            Random Forest Reg      &      &      &      &      &      &      &      &  &  0.98 & 0.91\\
            Extra Trees     &      &      &      &      &      &      &      &   &  & 0.91
\end{tabular}
\caption{Correlations between the base learners for the KNP fire.}
\label{tab:IX:base_correlation}
\end{table}

\begin{table}
\begin{tabular}{l|cccc}
Parameter & KNP & Windy & CZU & Kincade\\\hline
Elastic net & -0.08 (0.03) & -0.09 (0.03) & -0.05 (0.01) & -0.06 (0.02) \\
Decision Tree  & 2e-3 (4e-3) & -2e-3 (3e-3) & 4e-3 (4e-3) & -2e-4 (4e-3) \\
Ridge Reg & -0.02 (0.03) & 0.05 (0.04) & -0.11 (0.03) & 0.04 (0.02) \\
Lasso Reg & 0.05 (0.04) & -0.02 (0.05) & 0.14 (0.03) & -8e-3 (0.03) \\
KNeighbors Reg & -0.05 (7e-3) & -0.03 (6e-3) & -0.04 (7e-3) & 0.03 (7e-3) \\
Gradient Boosting Reg & 0.10 (0.01) & 0.13 (0.01) & 2e-3 (0.01) & 0.16 (0.01) \\
XGBoost Reg & 0.10 (0.01) & 0.08 (0.01) & 0.17 (0.01) & 0.07 (0.01) \\
Bagging Reg & 0.01 (0.03) & -0.03 (0.02) & -0.08 (0.02) & 1e-3 (0.03) \\
Random Forest Reg & 0.02 (0.03) & -1e-3 (0.02) & -0.12 (0.02) & -0.02 (0.03) \\
Extra Trees Reg & 0.56 (0.02) & 0.60 (0.01) & 0.94 (0.01) & 0.49 (0.02) \\
Multilayered Perceptron Reg & 0.29 (7e-3) & 0.20 (7e-3) & 0.18 (6e-3) & 0.29 (7e-3) \\
\end{tabular}
\caption{Fitted values (standard errors), separate by fire, for the regression coefficients $\beta_j$.
}
\label{tab:Fit:withinfire}
\end{table}

Methods are compared in Table \ref{tab:CV:withinfireCV} using prediction Root Mean Squared Error (RMSE), classification accuracy for all categories, and classification accuracy for high burn severity categories.  For all four fires the non-spatial linear regression model performs poorly with high RMSE and low classification accuracy.  Including either the SL model in the mean or spatial dependence gives  a large reduction in MSE and increase in classification accuracy.  Across the four fires, the classification accuracy of the spatial SL approach for predicting the burn severity category is between 58\% and 71\%, and thus the predictions are fairly reliable.

\begin{table}
\centering
\begin{tabular}{lll|cccc}
Fire & Mean & Covariance & RMSE & CA & CA-High \\\hline
KNP  & LR & Independent & 213 & 28.0 & 0.03 \\
       & LR & Spatial   & 115 & 57.5 & 57.9  \\
       & SL & Independent  & 121 & 55.7 & 53.6  \\
       & SL & Spatial    & 114 & 58.0 & 55.4  \\
&\vspace{-6pt}\\
Windy & LR & Independent & 215 & 32.1 & 9.8\\
       & LR & Spatial      & 98 & 66.0 & 78.1 \\
       & SL & Independent  & 112 & 61.6 & 73.2  \\
       & SL & Spatial    & 98 & 66.5 & 77.8  \\
&\vspace{-6pt}\\
CZU & LR & Independent & 207 & 37.3 & 39.0  \\
       & LR & Spatial      & 89 & 70.1 & 84.8  \\
       & SL & Independent  & 95 & 69.3 & 84.8 \\
       & SL & Spatial    & 91 & 70.7 & 86.3  \\
&\vspace{-6pt}\\
Kincade & LR & Independent & 164 & 43.0 & 17.5 \\
       & LR & Spatial      & 88 & 66.8 & 67.8\\
       & SL & Independent  & 94 & 65.3 & 69.1 \\
       & SL & Spatial    & 82 & 69.3 & 73.8  \\
\end{tabular}
\caption{Within-fire cross-validation root mean squared prediction error (RMSE), classification accuracy (CA) percent for fire severity level (high enhanced growth (dNBR from -500 - -251), low enhanced regrowth (-250 - -101), unburned (-100 - 99), low (100 - 269), moderate-low (270 - 439), moderate-high (440 - 659), and high (660 - 1300)) and CA for the high  categories for Linear Regression (LR) and Super Learner (SL) models with independent and spatial error structure. }
\label{tab:CV:withinfireCV}
\end{table}

\subsection{Combined-Fire Predictions}\label{s:app:combined_preds} 

In contrast to within-fire predictions, combined-fire predictions combine data from multiple individual fires into one dataset to form predictions. By combining the fires, we can identify how well the model fits differ across the entire region in California that these fires span. The fitted values of the parameters from the spatial SL method in Table \ref{tab:Fit:combinedfire} again show that Extra Trees Regression is the dominant learner, followed by Multilayer Perceptron Regression and XGBoost Regression.  Methods are compared in Table \ref{tab:CV:combinedfire}.  The spatial SL model has the lowest RMSE and highest overall classification accuracy, followed closely by the spatial linear regression model, which give the highest overall classification accuracy for the high category.  

\begin{table}
\centering
\begin{tabular}{l|c}
Parameter & Combined \\\hline
Elastic Net & -0.01 (0.01)\\
Decision Tree & -2e-3 (2e-3)\\
Ridge Reg & 6e-3 (0.01)\\
Lasso Reg & -0.02 (0.01)\\
K Neighbors Reg & -0.01 (3e-3)\\
Gradient Boosting Reg & 0.08 (6e-3)\\
XGBoost Reg & 0.16 (6e-3)\\
Bagging Reg & 0.03 (0.01)\\
Random Forest Reg & 0.04 (0.01)\\
Extra Trees Reg & 0.63 (8e-3)\\
Multilayered Perceptron Reg & 0.12 (4e-3)\\
\end{tabular}
\caption{Fitted values (standard errors), combining all fires, for the regression coefficients $\beta_j$.
}
\label{tab:Fit:combinedfire}
\end{table}

\begin{table}
\centering
\begin{tabular}{lll|ccc}
Fire & Mean & Covariance & RMSE & CA & CA-High \\\hline
Combined & LR & Independent & 223 & 29.0 & 8.26 \\
    & LR & Spatial & 100 & 64.5 & 76.54 \\
    & SL & Independent & 108 & 62.4 & 59.06\\
    & SL & Spatial & 97 & 65.7 & 62.98
\end{tabular}
\caption{Combined-fire cross-validation root mean squared prediction error (RMSE), classification accuracy (CA) percent for fire severity level (high enhanced growth (dNBR from -500 - -251), low enhanced regrowth (-250 - -101), unburned (-100 - 99), low (100 - 269), moderate-low (270 - 439), moderate-high (440 - 659), and high (660 - 1300)) and CA for the high  categories for Linear Regression (LR) and Super Learner (SL) models with independent and spatial error structure.  }
\label{tab:CV:combinedfire}
\end{table}

\subsection{Variable importance measures}\label{s:app:variableimportance}

While the spatial SL method provides solid prediction performance, it does not inherently provide measures of the effect of individual covariates, which is one of our main objectives.  Therefore, in this section we use interpretable ML tools to isolate the contribution of individual predictors on burn severity.  Due to the high performance of the Extra Trees Regressor at low computational cost, we opted to analyze variable importance using this model. A fitted attribute within the scikit-learn Python package \citep{scikit-learn} allowed us to extract the strongest predictors of wildland fire burn severity in California. The strongest predictors were calculated using Permutation Feature Importance (PFI; \cite{fisher2019all}).  The PFI algorithm ranks the importance of covariates as follows. First, a covariate column is selected within the dataset and its entries are shuffled 10 times. Then, the Extra Trees model is refit with the corrupted covariate column. Finally, a metric of importance is determined by evaluating the reduction of model score with the corrupted column, i.e., if shuffling PM SM resulted in a great reduction in model score, then that would indicate that the afternoon soil moisture is important for predicting burn severity in a given region. This process is repeated for each covariate of interest to determine a ranking. The top five strongest predictors for each individual fire are reported in Table \ref{tab:VI:withinfire}.  The leading predictors for within-fire prediction included the daily average dew point temperature (TDMEAN), normalized differenced vegetation index (NDVI), elevation of topography (Elevation), afternoon soil moisture (PM SM), morning soil moisture (AM SM), daily average air temperature (TMEAN), and vapor pressure maximum (VPDMAX).

\begin{table}
{\footnotesize \begin{tabular}{lc|lc|lc|lc}
KNP & VI & Windy & VI & CZU & VI & Kincade & VI \\\hline
TDMEAN (trend) & 0.23 & NDVI (current) & 0.33 & PM SM (current) & 0.38 & NDVI (current) & 0.54\\
NDVI (current) & 0.19 & Elevation & 0.17 & Elevation & 0.24 & Elevation & 0.15\\
Elevation & 0.14 & PM SM (trend) & 0.11 & NDVI (current) & 0.08 & AM SM (trend) & 0.07\\
PM SM (trend) & 0.10 & PM SM (current) & 0.08 &  TDMEAN (current) & 0.07 & VPDMAX (current) & 0.04 \\
AM SM (trend) & 0.07 & TDMEAN (current) & 0.07 & TMEAN (trend) & 0.07 & PM SM (trend) & 0.04\\
\end{tabular}}
\caption{Variable important (VI) measures for the Extra Trees learner on the four separate fires. Top variables include: daily average dew point temperature (TDMEAN), normalized differenced vegetation index (NDVI), elevation of topography (Elevation), afternoon soil moisture (PM SM), morning soil moisture (AM SM), daily average air temperature (TMEAN), vapor pressure maximum (VPDMAX).}
\label{tab:VI:withinfire}
\end{table}

\begin{figure}
\centering 
\includegraphics[width=0.7\textwidth]{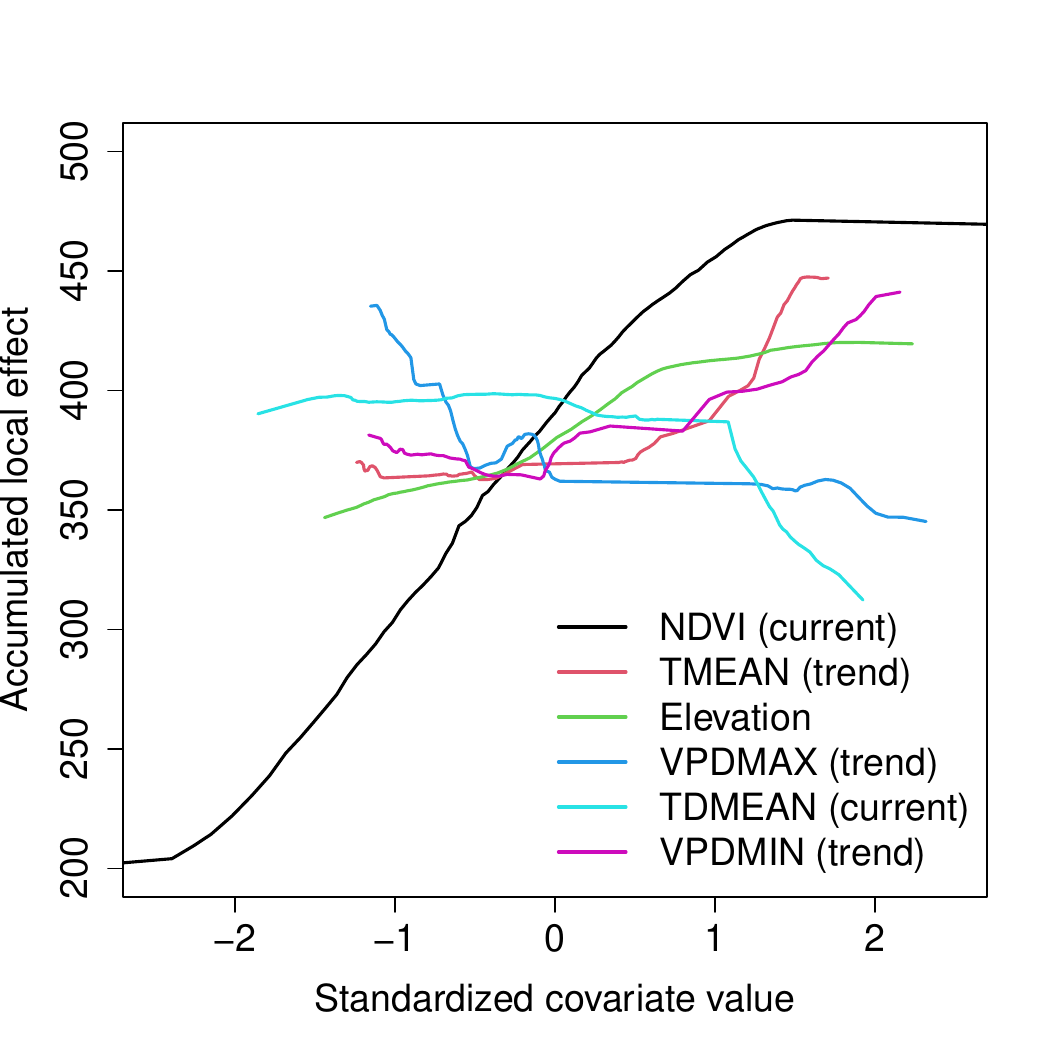}
\caption{Accumulated local effects of the  most important predictors.} \label{f:ME}
\end{figure}

The ranking of variable importance is similar for the combined analysis. The five most important variables are  NDVI (current) (VI = 0.34), elevation (0.20), PM SM (current) (0.20), LST (current) (0.10) and PM SM (trend) (0.08).  
In addition to ranking variables based on importance, Figure \ref{f:ME} plots estimates of the direction of the effects of these covariates to illustrate their relationship with burn severity.  The effects are estimated using the Accumulated Local Effects (ALE) of \cite{ale}. The local effect of a covariate is estimated by conditioning on all other covariates and computing the mean burn severity as estimated by the fitted SL model as a function of the covariate.  These local estimates are then averaged over the distribution of the other covariates.  The covariates with most variability in their ALE are plotted in Figure \ref{f:ME}. These plots suggest that the regions that are the most susceptible to burn damage are those with
high greenness (NDVI, current), increasing temperature (TMEAN, trend), high elevation, decreasing vapor pressure definite maximum (VPDMAX, trend), low dew point temperature (TDMEAN, current) and increasing vapor pressure deficit minimum (VPDMIN, trend).

\section{Discussion}\label{s:discussion}

This papers presents the results of evaluating the post-fire burn severity of four California wildfires between 2019-2021 using Super Learner regression and geostatistical techniques designed to handle large and complex datasets.  Although computationally expensive, the SL algorithm outperformed linear regression. This result is consistent with papers that have shown that ensemble learning reliably provides a more robust, accurate model than traditional machine learning algorithms alone. We built upon the work of \cite{Pascolini-Cambell2021ECOSTRESS} by including additional metrics to evaluate burn severity, such as those derived from the SMAP satellite. For the selected fires, hydrological data from ECOSTRESS proved less important than AM soil moisture as the leading predictor of burn severity. This aligns with \cite{Jensen} findings of the importance of pre-season
soil moisture as a strong predictor.  Future work could explore the importance of elevation, NDVI, and PM soil moisture for wildland fire prediction.

In summary, our mixed-model approach offers a compelling alternative to single learners, physics-based simulators, empirical models, and statistical models to predict wildland burn severity in California. By leveraging the spatial SL method, researchers can achieve accurate predictions of burn severity and better our understanding of the drivers of burn severity which is needed for pre-fire season monitoring. Identifying the most important variables to be mindful of, such as evening soil moisture, can informs Californians of ways to effectively allocate their time and resources to prepare for and respond to wildland fire incidences. In doing so, the loss of human life, property, resources, and ecosystems could be minimized.

\section*{Acknowledgements}
 
This work was carried out as part of a Research Experience for Undergraduates (REU). The authors would like to thank the organizers of the 2023 North Carolina State University “Directed Research for Undergraduates in Mathematics and Statistics” (DRUMS) program for their support and guidance. Funding was provided by the National Science Foundation (Grant DMS-2051010) and the National Security Agency (Grant H98230-23-1-0009). This work was also partially supported by National Science Foundation grant DMS-2152887. In addition, this research was partially carried out at the Jet Propulsion Laboratory, California Institute of Technology, under a contract with the National Aeronautics and Space Administration (80NM0018D0004). The authors would like to thank these collaborators  for their insight and resources. Furthermore, the authors acknowledge the Minnesota Supercomputing Institute (MSI) at the University of Minnesota for providing resources that contributed to the research results reported within this paper. Lastly, the authors would like to acknowledge the computing resources provided by North Carolina State University High Performance Computing Services Core Facility. 

\begin{appendices}

\section*{Appendix A.1: Training individual learners}\label{s:A1}

Part of the Super Learner algorithm is choosing the library of base learning models. These models were chosen to include a range of linear and non-linear methods. Specifically, the set comprised of Elastic Net, Decision Tree, Ridge Regression, Lasso Regression, K-nearest neighbors, Gradient Boosting, Extreme Gradient Boosting (XGBoost), Bagging, Random Forest, Extra Trees, and a custom-built Neural Network model. All these models were trained using their scikit-learn default hyperparameters, except where explicitly mentioned otherwise. Refer to the Super Learner code on our GitHub page (\href{https://github.com/Nicholas-Simafranca/Super_Learner_Wild_Fire.git}{https://github.com/Nicholas-Simafranca/Super\_Learner\_Wild\_Fire.git}) for specific hyperparameters used.

The Neural Network model was implemented in PyTorch, comprising four fully connected hidden layers. All hidden layers used a rectified linear unit (ReLU) activation function. The output layer was a single neuron as this is a regression task. The model was trained using a mean squared error loss function and the Adam optimizer. The number of training epochs was set to 200 with a batch size of 115 and a learning rate of 0.01.

To generate the the out-of-fold predictions, a 10-fold cross-validation was performed on the entire dataset. In each fold of this cross-validation, each base model was trained on 80 percent of the data, and predictions were generated on the remaining 20 percent. This process was repeated such that every sample in the dataset had an associated set of out-of-fold predictions, one from each base learner. The out-of-fold predictions from all the base learners were then stacked horizontally to form a new matrix of meta-features, denoted as the $Z$ matrix, one for each base learner (Figure \ref{f:diagram}). Along with these meta-features, dNBR and corresponding latitude and longitude were also stored. The process of generating out-of-fold predictions ensured that the base learners and the meta-model remained decoupled, preventing data leakage and ensuring robustness of the ensemble.

In addition to generating out-of-fold predictions, the base learners were also trained on the entire training dataset, and predictions were made on the same dataset to obtain a set of fitted values. These fitted values were used to calculate the training RMSE for each base learner.

\section*{Appendix A.2: Fitting the spatial model}\label{s:A2}

Once all base learners had been trained and their out-of-fold predictions had been generated, the meta-model is trained using the {\tt GpGp}: Fast Gaussian Process Computing package in R \citep{GpGp}. This package supports spatial models, including our exponential isotropic covariance function, with many ways of increasing computational efficiency. It implements Vecchia's Gaussian approximation, which is one of the most efficient Gaussian process approximators \citep{GuinnessVecchias}. Furthermore, Guinness (2021) improves upon Vecchia's approximation in the package by implementing a Fisher scoring algorithm for efficient computing of the maximum likelihood estimation of parameters, $\beta, \phi, \tau^2, \sigma^2$. It also uses Vecchia's Gaussian approximation to predict at new unsampled locations, a form of Kriging \citep{DenisSpatialStats}. This utilizes computing the inverse Cholesky factorization of the covariance matrix when finding the conditional expectation as described in \cite{DenisSpatialStats} and \cite{KatzfussVecchias}. In its prediction, the package also orders and groups observations based on spatial proximity, further decreasing computing time \cite{DenisSpatialStats}. 




\end{appendices}


\bibliography{sn-bibliography}

\end{document}